\documentclass[letterpaper]{article} % DO NOT CHANGE THIS
\usepackage{aaai24}  % DO NOT CHANGE THIS
\usepackage{times}  % DO NOT CHANGE THIS
\usepackage{helvet}  % DO NOT CHANGE THIS
\usepackage{courier}  % DO NOT CHANGE THIS
\usepackage[hyphens]{url}  % DO NOT CHANGE THIS
\usepackage{graphicx} % DO NOT CHANGE THIS
\usepackage{natbib}  % DO NOT CHANGE THIS AND DO NOT ADD ANY OPTIONS TO IT
\usepackage{caption} % DO NOT CHANGE THIS AND DO NOT ADD ANY OPTIONS TO IT
\usepackage{algorithm}
\usepackage{algorithmic}

\usepackage{amsmath}
\usepackage{bm}
\usepackage{amssymb}
\usepackage{multirow}
\usepackage{graphicx}
\usepackage{subfigure}
\usepackage{capt-of}
\usepackage{booktabs}

\usepackage{newfloat}
\usepackage{listings}
\DeclareCaptionStyle{ruled}{labelfont=normalfont,labelsep=colon,strut=off} % DO NOT CHANGE THIS
\floatstyle{ruled}
\newfloat{listing}{tb}{lst}{}
\floatname{listing}{Listing}
\title{Liberating Seen Classes: Boosting Few-Shot and Zero-Shot Text Classification\\ via Anchor Generation and Classification Reframing}
\author {
    Han Liu\textsuperscript{\rm 1},
    Siyang Zhao\textsuperscript{\rm 1},
    Xiaotong Zhang\textsuperscript{\rm 1}\thanks{Corresponding author.},
    Feng Zhang\textsuperscript{\rm 2},
    Wei Wang\textsuperscript{\rm 3},\\
    Fenglong Ma\textsuperscript{\rm 4},
    Hongyang Chen\textsuperscript{\rm 5},
    Hong Yu\textsuperscript{\rm 1},
    Xianchao Zhang\textsuperscript{\rm 1}
}
\affiliations {
    \textsuperscript{\rm 1} Dalian University of Technology, Dalian, China\\
    \textsuperscript{\rm 2} Peking University, Beijing, China\\
    \textsuperscript{\rm 3} Shenzhen MSU-BIT University, Shenzhen, China\\
    \textsuperscript{\rm 4} The Pennsylvania State University, Pennsylvania, USA\\
    \textsuperscript{\rm 5} Zhejiang Lab, Hangzhou, China\\
    liu.han.dut@gmail.com, zhao\_siyang@mail.dlut.edu.cn, zxt.dut@hotmail.com, zfeng.maria@gmail.com,\\ ehomewang@ieee.org,
    fenglong@psu.edu, dr.h.chen@ieee.org, \{hongyu, xczhang\}@dlut.edu.cn
}

\usepackage{bibentry}
\begin{document}

\maketitle

\begin{abstract}
Few-shot and zero-shot text classification aim to recognize samples from novel classes with limited labeled samples or no labeled samples at all. While prevailing methods have shown promising performance via transferring knowledge from seen classes to unseen classes, they are still limited by (1) Inherent dissimilarities among classes make the transformation of features learned from seen classes to unseen classes both difficult and inefficient. (2) Rare labeled novel samples usually cannot provide enough supervision signals to enable the model to adjust from the source distribution to the target distribution, especially for complicated scenarios. To alleviate the above issues, we propose a simple and effective strategy for few-shot and zero-shot text classification. We aim to liberate the model from the confines of seen classes, thereby enabling it to predict unseen categories without the necessity of training on seen classes. Specifically, for mining more related unseen category knowledge, we utilize a large pre-trained language model to generate pseudo novel samples, and select the most representative ones as category anchors. After that, we convert the multi-class classification task into a binary classification task and use the similarities of query-anchor pairs for prediction to fully leverage the limited supervision signals. Extensive experiments on six widely used public datasets show that our proposed method can outperform other strong baselines significantly in few-shot and zero-shot tasks, even without using any seen class samples.
\end{abstract}

 \begin{figure*}[t]
 	\centering
 	\includegraphics[width=1\textwidth]{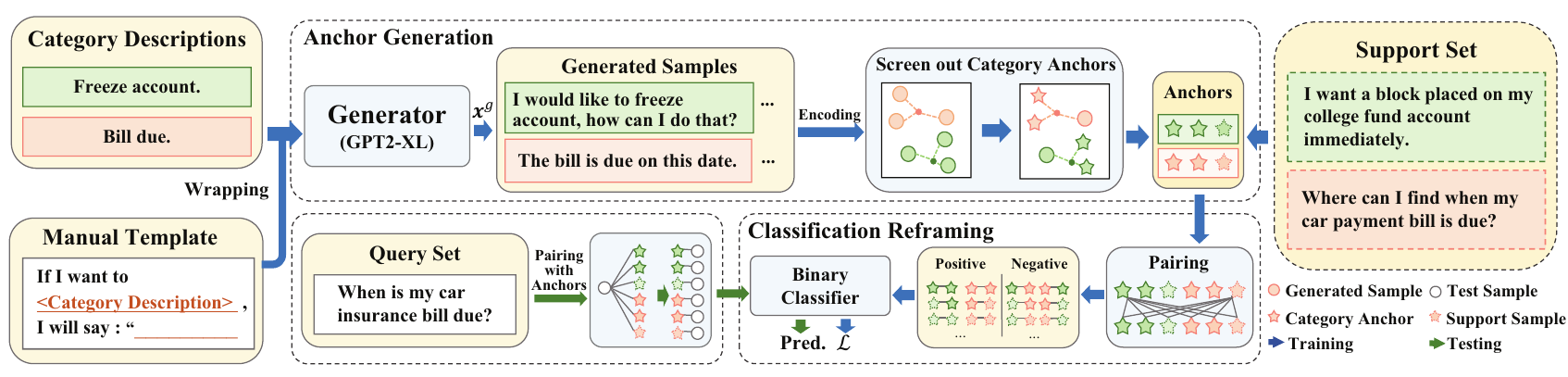}
 	\caption{The framework of our proposed method.}
 \label{framework}
 \end{figure*}

\section{Introduction}
Text classification is widely used in various real applications, such as intent detection, news classification and so on \cite{DBLP:journals/nca/KumarA21}. This fundamental task aims to assign predefined labels or categories to the given texts. Conventional text classification approaches typically demand copious amounts of labeled training data for achieving high accuracy \cite{DBLP:conf/acl/JohnsonZ17, DBLP:conf/naacl/DevlinCLT19}. However, the acquisition of such labeled data can be a challenging and expensive endeavor in many real-world scenarios, thereby prompting the emergence of few-shot and zero-shot text classification.

Few-shot text classification (FSTC) and zero-shot text classification (ZSTC) methods are dedicated to addressing these challenges by obtaining powerful classifiers using a sparse set of labeled samples, or even in cases where no labeled samples are available. FSTC focuses on constructing classifiers from a small number of labeled samples, while ZSTC endeavors to classify texts into classes that have not been encountered during the training phase.

For few-shot text classification, meta-learning methods help the model quickly adapt to novel classes by constructing training tasks that mimic the testing tasks from seen classes \cite{finn2017model,snell2017prototypical, DBLP:conf/aaai/0008Z0ZMWC0Z23}. Fine-tuning methods pre-train the model on the base set and adapt it to unseen classes by fine-tuning a part of parameters with support samples \cite{howard-ruder-2018-universal,gururangan-etal-2020-dont,DBLP:conf/emnlp/ZhangHLWWYSX20}. Recently, prompt-based methods efficiently utilize pre-trained models and achieve great performance \cite{lmbff,efl}. For zero-shot text classification, many reliable methods have emerged in intent recognition, which can be roughly divided into transformation-based and projection-based methods. The former methods transform the semantic space of seen classes into unseen classes through the similarity matrix \cite{xia2018zero,DBLP:conf/emnlp/LiuZFFLWL19}, and the latter methods calculate the similarity between labels and samples from unseen classes by mapping them into the same space \cite{chen2016zero,kumar2017zero}. Some generation methods are also applied to expand training data in zero-shot text classification \cite{DBLP:conf/emnlp/SchickS21a,DBLP:conf/emnlp/YeGLXF00K22}, but due to the limitation of generation quality, most of them are used for simple tasks like binary-classification or three-classification rather than complex tasks.

After analyzing previous works comprehensively, we find that whether employing few-shot or zero-shot methods, conventional approaches strive to transfer knowledge from seen classes to unseen classes, which leads to two unavoidable problems. First, it is hard to properly adapt the model trained on seen classes to unseen classes simply using supervision signals from few labeled novel samples. That is because the inherent dissimilarities among classes introduce difficulty and inefficiency in transferring learned features from seen to unseen classes. What's more, the lack of labeled samples for novel classes typically fails to offer sufficient guidance for the model to smoothly transition from the source distribution to the target distribution, especially in complex scenarios. In response to the issues outlined above, in this paper, we propose an effective method to break out the limitations of seen classes, where we only use a small set of generated samples as category anchors for each unseen class, yielding substantial performance improvements. The framework of our approach is shown in Figure \ref{framework}. In pursuit of better understanding of the target categories, we employ a pre-trained language generation model to generate pseudo unseen class data with wrapped class descriptions. Subsequently, we refine this generation process by identifying and retaining the most representative samples, establishing them as effective class anchors. Specifically, we compute prototypes for unseen classes and sample class anchors based on the distance between generated instances and the prototypes. In order to fully harness the potential of the generated supervisory signals and enhance performance, we strategically reframe the complex multi-class classification task into a binary classification framework by constructing query-anchor pairs.

\section{Related Work}
\subsection{Few-Shot Text Classification}
Few-shot text classification (FSTC) aims to recognize novel classes with very few labeled samples. Existing FSTC methods can be broadly categorized into two types: meta-learning methods and fine-tuning methods. Meta-learning methods help the model quickly adapt to novel classes by making training tasks that mimic the testing tasks from seen classes, which can be further subdivided into optimization-based and measure-based methods. Optimization-based meta-learning methods typically involve multiple rounds of gradient updates to learn a set of meta-parameters that can be quickly adapted to new tasks with a few labeled samples, such as MAML \cite{finn2017model}. Measure-based meta-learning methods aim to learn a metric space where similar samples are close to each other, and dissimilar samples are far apart. One popular approach is the prototypical network \cite{snell2017prototypical}, which learns an embedding space where each class is represented by its prototype and classifies samples based on the similarity between the samples and the prototypes. 

Fine-tuning methods pre-train the model on the base set and adapt it to new classes by fine-tuning a part of parameters with support samples in unseen classes \cite{howard-ruder-2018-universal,gururangan-etal-2020-dont}. DNNC \cite{DBLP:conf/emnlp/ZhangHLWWYSX20} pre-trains the model based on the large-scale text inference datasets, and converts the few-shot classification task into a natural language inference task. Recently, prompt-based methods efficiently utilize pre-trained models and achieve great performance \cite{lmbff,efl}. Prompt-based methods provide specificity in the form of input prompts and guide the model to generate desired responses for targeted tasks.

\subsection{Zero-Shot Text Classification}
Zero-shot text classification (ZSTC) deals with novel classes without annotated training data. 
Notably, a multitude of mature methods have emerged, particularly in the domain of intent recognition, which can be broadly categorized into two approaches: transformation-based and projection-based methods. The transformation-based methods \cite{xia2018zero,DBLP:conf/emnlp/LiuZFFLWL19}
create a similarity matrix with the embeddings of label descriptions. This matrix is employed to transfer the prediction vectors from seen classes to unseen classes. The basic idea is to establish a connection between seen classes and unseen classes using the semantic relationships captured in the similarity matrix. On the other hand, projection-based methods
\cite{chen2016zero,kumar2017zero} aim to project both the label descriptions and utterances into a shared semantic space. By doing so, the similarity between the labels and the utterances can be calculated, enabling effective classification of unseen classes. Several generation methods are employed to expand training data in ZSTC \cite{DBLP:conf/emnlp/SchickS21a,DBLP:conf/emnlp/YeGLXF00K22}. However, because of the constraints of generation quality, they are predominantly utilized for straightforward tasks such as binary/three-classification.

\section{Preliminaries}
\subsubsection{Few-Shot Learning}
In this paper, we address the problem of few-shot text classification under the $N$-way $K$-shot setting \cite{cvpr/SungYZXTH18}. In this setting, each task consists of $N$ classes, and each class has $K$ labeled samples as supports. The main objective is to predict the class label for a given query. Traditional FSTC methods aim to transfer knowledge from adequately labeled samples of seen classes $\mathcal{C}_\text{seen}$ to unseen classes $\mathcal{C}_\text{unseen}$, where $\mathcal{C}_\text{seen} \cap \mathcal{C}_\text{unseen} = \emptyset$. As a result, the presence of seen classes becomes necessary for achieving promising performance in these methods. However, in this paper, we propose an effective method that distinctly eliminates the necessity for seen classes, thereby we introduce a pure few-shot learning setting that only utilizes the $N \times K$ samples (i.e., the support set) and the class descriptions of $\mathcal{C}_\text{unseen}$ for prediction. And we do not rely on any seen classes. By adopting this approach, we aim to demonstrate the effectiveness of our proposed method in handling few-shot text classification without the need for seen class information. Notice that our setting can be smoothly transferred to ZSTC tasks with $N$ novel classes ($N$-way $0$-shot tasks), which is an extreme case of few-shot text classification. 

\subsubsection{Zero-Shot Learning}
Traditional zero-shot learning endeavors to predict the label $y_{\text{te}}\in\mathcal{C}_{\text{unseen}}$ of a test sample $\boldsymbol{x}_{\text{te}}$ by transferring knowledge from seen classes \cite{chen2016zero}. This makes the model performance highly dependent on seen class. In this paper, we propose a novel zero-shot method without relying on any seen data for knowledge transfer, which performs the classification task solely based on the information extracted from unseen class descriptions and pre-trained language models. That means, under the settings of our method, only description information for the unseen classes $\mathcal{C}_{\text{unseen} }$ is available.

\section{The Proposed Method}
Our proposed method is designed to tackle the challenge of few-shot and zero-shot tasks without the need for any seen classes. It encompasses two primary parts: anchor generation and classification reframing.

\begin{table}[t]
	\small
	\centering
	\tabcolsep=0.1cm
	\begin{tabular}{l|l}
		\toprule
  \multicolumn{1}{l|}{\textbf{Symbol}} & \multicolumn{1}{c}{\textbf{Explanation}} \\
  \midrule
		$\boldsymbol{\hat{y}}$      &  The class description of the class $y$. \\
		$\text{T}(\boldsymbol{\hat{y}})$      &  The instruction of the class $y$. \\
		$\boldsymbol{x}^{g}$      
  &  A generated sample. \\
		$\boldsymbol{e}$      & The embedding vector of data. \\
		$\boldsymbol{p}_{y}$      & The generation prototype of the class $y$. \\
		${\text{X}}_{y}$
		& The anchor set of the class $y$.\\
		${\mathbb{X}}$
		& The anchor set for all unseen classes. \\
		$\boldsymbol{v}({\boldsymbol{x}}_{i},{\boldsymbol{x}}_{j}) $    
		&The similarity indicator vector between $\boldsymbol{x}_{i}$ and $\boldsymbol{x}_{j}$. \\
  $\hat{v}$ & A similarity score.\\
		\bottomrule
	\end{tabular}
 	\caption{The symbol definition.}
	\label{tab:symbol}
\end{table}

\subsection{Anchor Generation}
\subsubsection{Data Generation via Category Descriptions}
Pre-trained language models (PLMs) have demonstrated exceptional capabilities in generating high-quality text samples, leading to significant advancements in emotion recognition and natural language inference tasks \cite{DBLP:conf/emnlp/SchickS21a,DBLP:conf/emnlp/YeGLXF00K22}. In this paper, we extend the application of PLMs to address the challenges of few-shot and zero-shot text multi-classification tasks.

Given an unseen class $y\in\mathcal{C}_\text{unseen}$, we possess valuable category descriptions $\boldsymbol{\hat{y}}$, which serve as the foundation for constructing label-specific instruction $\text{T}(\boldsymbol{\hat{y}})$. The manual template $\text{T}(\cdot)$ is designed to encapsulate the category descriptions and streamline the generation of pseudo unseen class data. To achieve this, we employ a one-way generator. Specifically, given a class description $\boldsymbol{\hat{y}}$ for the category $y$, we utilize the instruction $\text{T}(\boldsymbol{\hat{y}})$ to continuously sample tokens from the PLM in a sequential manner, resulting in a sequence of tokens denoted as $\boldsymbol{x}^{g} = \{x_1, x_2, ..., x_{k-1}\}$. The generation process is formulated as follows:
\begin{eqnarray}
x_{k} \sim p_{\text{PLM}}(x_{k}|\text{T}(\boldsymbol{\hat{y}}),x_{1},...,x_{k-1}),
\end{eqnarray}
where $k$ starts from $1$ and continues until a specific marker is reached. To ensure that $\boldsymbol{x}^{g}$ ends in a reasonable position, we adopt the "quotation-mark-end" strategy \cite{DBLP:conf/emnlp/SchickS21a}. Specifically, we design the instruction $\text{T}(\boldsymbol{\hat{y}})$ to conclude with an opening quotation mark, and employ the first generated quotation mark as the end marker. In order to enhance the diversity of the generated samples $\boldsymbol{x}^{g}$, we utilize the top-p and top-k sampling techniques \cite{DBLP:conf/iclr/HoltzmanBDFC20,DBLP:conf/acl/LewisDF18,DBLP:conf/acl/ChoiBGHBF18}. These strategies effectively contribute to the production of varied and informative pseudo unseen class data, which improves the performance of our approach in few-shot and zero-shot tasks.

\subsubsection{Screen out Category Anchors}
In order to create precise representations for the novel classes, we extract a set of category anchors that are close to the class prototypes. Specifically, given a generated set $\{ \boldsymbol{x}^{g}_{y1}, \boldsymbol{x}^{g}_{y2}, ..., \boldsymbol{x}^{g}_{yn} \}$ comprising $n$ generated texts of the category $y$, our approach utilizes a pre-trained encoder to derive $d$-dimensional semantic embedding vectors $\boldsymbol{e}$ for each generated sample $\boldsymbol{x}^{g}$,
\begin{eqnarray}
\boldsymbol{e}= \text{Encoder}(\boldsymbol{x}^{g})\in \mathbb{R}^{d}.
\end{eqnarray}

For cases where the encoder is BERT \cite{DBLP:conf/naacl/DevlinCLT19}, we extract the output embedding $\boldsymbol{e}$ at the special token $\texttt{[CLS]}$ as the semantic embedding of $\boldsymbol{x}^{g}$. For each tested class $y\in \mathcal{C}_\text{unseen}$, we obtain the semantic embedding set $\{ \boldsymbol{e}_{y1},...,\boldsymbol{e}_{yn}\}$ for the generated samples and calculate the average vector to form the class prototype $\boldsymbol{p}_{y}\in \mathbb{R}^{d}$. Finally, to create a set of category anchors $\text{X}_{y}=\{ \boldsymbol{x}^{g}_{y1},...,\boldsymbol{x}^{g}_{yP} \}$ for unseen class $y$, we sample $P$ generated texts from the probability distribution based on the Euclidean distance between each embedding $\boldsymbol{e}$ and the corresponding prototype $\boldsymbol{p}_{y}$.

In the few-shot setting, where each unseen class has a few labeled samples as support samples, we directly integrate these valuable samples into the corresponding category's anchor set.
More precisely, for $N$-way $K$-shot tasks, the anchor set for each class is constructed as $\text{X}_{y}=\{ \boldsymbol{x}^{g}_{y1},...,\boldsymbol{x}^{g}_{yP},\boldsymbol{x}_{y1},..., \boldsymbol{x}_{yK}\}$, where $\{ \boldsymbol{x}^{g}_{y1},...,\boldsymbol{x}^{g}_{yP}\}$ are the $P$ generated samples and $\{ \boldsymbol{x}_{y1},...,\boldsymbol{x}_{yK}\}$ are the $K$ support samples for the category. By repeating this process for all $N$ unseen classes, we obtain a total of $N \times (P+K)$ anchors.

\subsection{Classification Reframing}
It is natural that when two samples display a considerable degree of resemblance, they are likely in the same category   \cite{DBLP:conf/icsca/MukaharR21, DBLP:conf/nips/SimardCD92}. And transforming multi-class classification tasks into binary classification tasks could potentially enhance the model's performance, particularly under resource constraints \cite{DBLP:conf/emnlp/ZhangHLWWYSX20,DBLP:conf/icsca/MukaharR21, DBLP:conf/nips/SimardCD92}. Building upon the aforementioned concept, we proceed to reframe the intricate multi-class classification task as a binary classification challenge. We establish connections between pairs of anchor points, classifying them as either positive or negative pairs contingent upon their category affiliation.

In a more detailed manner, given the anchor set $\mathbb{X}=\text{X}_{1}\cup...\cup\text{X}_{N}$, where each $\text{X}$ corresponds to an unseen class. For any pair of anchors $\boldsymbol{x}_{i}, \boldsymbol{x}_{j}\in \mathbb{X}$, we categorize them as either positive or negative pairs, depending on whether they belong to the same category. To represent the similarity between two anchors, we introduce a 2-dimensional vector as the target similarity indicator $\phi ( \boldsymbol{x}_{i}, \boldsymbol{x}_{j}) = [m, 1-m] $, with $m$ defined as follows:
\begin{eqnarray}
	\label{phi}
	m = \left\{
	\begin{aligned}
		1 & & {y_i = y_j , } \\
		0 & & {y_i\neq y_j ,} \\
	\end{aligned}
	\right.
\end{eqnarray}
where $y_i$ and $y_j$ are the labels of $\boldsymbol{x}_{i}$ and $ \boldsymbol{x}_{j} $ respectively. By employing this scheme, we obtain $(N\times P)^2$  (in zero-shot tasks) or $(N\times (P+K))^2$  (in few-shot tasks) pairs for $N$ unseen classes, forming a new training set $\{((\boldsymbol{x}_{i},\boldsymbol{x}_{j}),\phi_{ij})\}$. To assess the similarity between two anchors, we use $\{ \texttt{[CLS]},\boldsymbol{x}_{i},\texttt{[SEP]},\boldsymbol{x}_{j},\texttt{[SEP]}\}$ as input and train a binary classifier based on BERT, which outputs a 2-dimensional vector $\boldsymbol{v} $:
\begin{eqnarray}
    \boldsymbol{v}(\boldsymbol{x}_{i},\boldsymbol{x}_{j}) = \text{softmax}(\boldsymbol{W}_{1}  \text{BERT} (\boldsymbol{x}_{i}, \boldsymbol{x}_{j}) + \boldsymbol{b}_{1}),
\end{eqnarray}
where $\boldsymbol{W}_1$ and $\boldsymbol{b}_{1}$ are learnable parameters. To guide the training, we apply label smoothing \cite{DBLP:conf/cvpr/SzegedyVISW16} and minimize the following loss function between the similarity vector and the target indicator, and $l$ represents the $l$-th element in the vector:
\begin{eqnarray}
\label{loss}
    \mathcal{L} = \frac{1}{|\mathbb{X}|^2}\sum\limits_{\boldsymbol{x}_{i},\boldsymbol{x}_{j}\in {\mathbb{X}}}\sum\limits_{l\in\{1,2\}} \boldsymbol{v}(\boldsymbol{x}_{i}, \boldsymbol{x}_{j})_l\text{log}\frac{\boldsymbol{v}({\boldsymbol{x}}_{i},{\boldsymbol{x}}_{j})_l}{\phi({\boldsymbol{x}}_{i},{\boldsymbol{x}}_{j})_l }.
\end{eqnarray}

By using classification reframing, we effectively convert the complex multi-class classification task into a binary classification problem. This strategic reformulation not only simplifies the training complexity under low-resource conditions but also adeptly captures class differences, amplifying the discriminative power of the model.

\subsection{Testing}
In the testing phase, given a test sample $\boldsymbol{x}^*$ from any unseen class, we follow the same procedure as in anchor pairs construction to form query-anchor pairs $(\boldsymbol{x}^*,\boldsymbol{x})$. Specifically, we utilize the binary classifier to obtain the similarity vector set $\text{V}=\{\boldsymbol{v}_1,...,\boldsymbol{v}_n\}$ between $\boldsymbol{x}^*$ and $\boldsymbol{x}$, where $n = P\times N$ denotes the total number of anchors, $P$ is the number of anchors per class and $N$ is the number of unseen classes in the test set. For each $\boldsymbol{v}_i=[v_{i}^0,v_{i}^1]$, we define $ \hat{v}_{i}=v_{i}^0-v_{i}^{1} \in \mathbb{R}^{1}$ as the similarity score, which is positively correlated with the probability that sample $\boldsymbol{x}^*$ shares the same label with anchor $\boldsymbol{x}_i$. Based on the computed similarity scores, we propose two simple prediction methods for the classification of the test sample.

\subsubsection{Predict by the Top-One Score}
We directly select the category associated with the anchor that exhibits the highest correlation with $\boldsymbol{x}^*$:
\begin{eqnarray}
y^* = \text{argmax}_{k}(\hat{v}_{i}), \boldsymbol{x}_i \in \text{X}_{k}.
\end{eqnarray}

\subsubsection{Predict by the Average Score}
We compute the prediction by averaging the similarity scores for each class as follows:
\begin{eqnarray}
    y^* =  \text{argmax}_{k}(\frac{1}{|\text{X}_{k}|}\sum\limits_{\boldsymbol{x}_i \in \text{X}_{k}} \hat{v}_{i}).
\end{eqnarray}

These methods provide straightforward and effective ways to predict the class label for the test sample $\boldsymbol{x}^*$ based on its similarity with the anchors, which further enhances the overall performance of our approach in multi-class text classification tasks. 

\section{Experiments}

\begin{table}[t]
\centering
\small
\renewcommand\arraystretch{1}{
    \begin{tabular}{lccc}
	\toprule
	\textbf{Dataset} & \textbf{Samples} & \textbf{Avg. Length} & \textbf{Train / Valid / Test} \\ \midrule
  	20News & 18828 & 279.32 & 8 / 5 / 7  \\
        Amazon & 24000 & 143.46 & 10 / 5 / 9 \\
	HuffPost & 36900 & 11.48 & 20 / 5 / 16  \\
	Reuters & 620 & 181.41 & 15 / 5 / 11 \\
        \midrule
	SNIPS & 13802  & 9.05 & 5 / 0 / 2 \\
	CLINC & 9000 & 8.34 & 50 / 0 / 10 \\
	\bottomrule
    \end{tabular}}
    \caption{Dataset statistics.}
    \label{statistics}
\end{table}

\begin{table*}[t]
    \small
    \centering
    \setlength{\tabcolsep}{3mm}{
    \begin{tabular}{l|cccc|cccc}
	\toprule
	{} & \multicolumn{4}{c|}{\textbf{20News}} & \multicolumn{4}{c}{\textbf{Amazon}}\\
        \cmidrule(lr){2-5} \cmidrule(lr){6-9} 
        {\textbf{Method}} & \multicolumn{2}{c|}{\textbf{1-shot}} & \multicolumn{2}{c|}{\textbf{5-shot}} & \multicolumn{2}{c|}{\textbf{1-shot}} & \multicolumn{2}{c}{\textbf{5-shot}}\\
	\cmidrule(lr){2-5} \cmidrule(lr){6-9} 
	{} & Acc & F1 & Acc & F1 & Acc & F1 & Acc & F1\\
	\midrule
        NN &  $0.3730$ & $0.3244$ & $0.5323$ & $0.5173$ & $0.4853$ & $0.4552$ & $0.6992$ & $0.6952$  \\
        Prototypical Network \quad\quad & $0.4721$ & $0.4426$ & $0.6014$ & $0.5948$   &$0.5407$ & $0.5320$ &  $0.6982$& $0.6979$\\
  	MAML  & $0.4309$ & $0.4042$ & $0.6194$ & $0.6095$ &$0.4900$& $0.4746$  & $0.6987$  & $0.6945$ \\
	Induction Network &$0.4138 $& $0.3715$ & $0.4922$ & $0.4910$ &  $0.5222$  & $0.5049$  & $0.5646$ & $0.5508$   \\
        TPN  & $0.5898$  & $0.5247$ & $0.7115$  & $0.6700 $  & $0.7072$  & $0.6745$ & $0.7672$  & $0.7330$    \\
        DS-FSL & $0.6019$ & $0.5729$ & $0.8064$ & $0.8011$    & $0.6592$ & $0.6434$ & $0.8454$ & $0.8422$ \\
        MLADA  & $0.6040$ & $0.5776$ & $0.7794$ & $0.7752$  &  $0.6328$ &$0.6143$ & $0.8234$  & $0.8184$\\ 
        ContrastNet   & $0.7229$   & $0.7047$  & $\bf{0.8418}$  & $\bf{0.8330}$ & $0.7606$ & $0.7548$ & $0.8442$ & $0.8369$\\
        \midrule
        \textbf{Ours-top}  & $\bf{0.7310}$ & $\bf{0.7267}$ & $\bf{0.8288}$ & $\bf{0.8284}$  & $\bf{0.7888}$  & $\bf{0.7814}$ & $\bf{0.8538}$  & $\bf{0.8518}$ \\
        \textbf{Ours-avg} & $\bf{0.7334}$  & $\bf{0.7276}$ & $0.8138$  & $0.8140$   & $\bf{0.7931}$  & $\bf{0.7870}$ & $\bf{0.8526}$  & $\bf{0.8510} $      \\
	\toprule
	{} & \multicolumn{4}{c|}{\textbf{HuffPost}} & \multicolumn{4}{c}{\textbf{Reuters}} \\
        \cmidrule(lr){2-5} \cmidrule(lr){6-9} 
        {\textbf{Method}} & \multicolumn{2}{c|}{\textbf{1-shot}} & \multicolumn{2}{c|}{\textbf{5-shot}} & \multicolumn{2}{c|}{\textbf{1-shot}} & \multicolumn{2}{c}{\textbf{5-shot}}\\
	\cmidrule(lr){2-5} \cmidrule(lr){6-9} 
	{}  & Acc & F1 & Acc & F1 & Acc & F1 & Acc & F1\\
	\midrule
        NN & $0.3092$ & $0.2440$ & $0.3949$ & $0.3725$ & $0.5431$ & $0.5154$ & $0.8018$ & $0.7980$ \\
   	Prototypical Network & $ 0.3436$ & $0.3159$ & $0.4940$ & $0.4889$ &$0.6233$ & $0.6075$ & $0.7188$ & $0.7117$ \\
  	MAML & $0.2453$  & $0.2349$ & $0.3374$ & $0.3284$ & $0.5394$ & $0.5275$ & $0.7942$ & $0.7900$ \\
	Induction Network  &  $0.3525$ & $0.3336$ & $0.4341$ & $0.4232$  & $0.5578$ & $0.5277$ & $0.6425$ & $0.6210$ \\
        TPN &  $0.500  5$  & $0.4780$ & $0.6665$ & $0.6569$ & $0.8368$ & $0.8022$ & $0.8614$ & $0.8291$ \\
        DS-FSL & $0.3946$ & $0.3759$ & $0.5986$ & $0.5892$ & $0.7572$ & $0.7447$ & $0.9404$ & $0.9392$ \\
        MLADA & $0.4433$ & $0.4188$ & $0.5736$ & $0.5650$ & $0.7178$ & $0.7028$ & $0.8252$ & $0.8215$ \\ 
        ContrastNet & $0.5045$ & $0.4909$ & $0.6718$ & $0.6621$ & $0.8787$ & $0.8699$ & $0.9443$ & $0.9417$ \\
        \midrule
        \textbf{Ours-top} & $\bf{0.6914}$ & $\bf{0.6875}$ & $\bf{0.7134}$ & $\bf{0.7106}$ & $\bf{0.9289}$ & $\bf{0.9263}$  & $\bf{0.9613}$ & $\bf{0.9612}$ \\
        \textbf{Ours-avg} & $\bf{0.6979} $ & $\bf{0.6945}$  & $\bf{0.7178}$  & $\bf{0.7146}$ & $\bf{0.9253}$ & 
    $\bf{0.9229}$ & $\bf{0.9605}$ & $\bf{0.9604}$ \\
	\bottomrule
	\end{tabular}}
    \caption{Experimental results of few-shot classification.}
    \label{tab:result1}
\end{table*}

\subsection{Datasets}
We conduct our experiments across four varied news and review classification datasets: 20News, Amazon, HuffPost, and Reuters, focusing on few-shot tasks in accordance with the experimental configuration outlined in \cite{ContrastNet}. Additionally, we follow  \cite{DBLP:conf/ijcai/SiL00LW21} to evaluate our method on intent classification datasets: SNIPS \cite{snips} and CLINC \cite{oos} on zero-shot tasks. Note that the training and validation classes are only used in baselines. The detailed statistics for all datasets are summarized in Table \ref{statistics}. 

\textbf{(1) 20News} \cite{20news} consists of 18828 documents collected from news discussion forums, categorized into 20 distinct topics.

\textbf{(2) Amazon} \cite{amazon} encompasses a vast collection of 142.8 million customer reviews spanning 24 different product categories. To maintain manageable dataset sizes, we follow the approach in \cite{mlada} and use a subset containing 1000 reviews per category.

\textbf{(3) HuffPost} \cite{iclr/BaoWCB20} comprises news headlines with 41 distinct classes, extracted from HuffPost articles published between 2012 and 2018. 

\textbf{(4) Reuters} \cite{iclr/BaoWCB20} consists of shorter articles from the Reuters news agency published in 1987. Similar to \cite{iclr/BaoWCB20}, we filter out multi-label articles and focus on 31 classes with a minimum of 20 articles per class.

\textbf{(5) SNIPS} \cite{snips} is a corpus specifically designed to evaluate the performance of voice assistants. It consists of 5 seen intents and 2 unseen intents. 

\textbf{(6) CLINC} \cite{oos} is an intent detection dataset which includes both in-scope and out-of-scope queries. It comprises 22,500 in-scope queries. We follow \cite{DBLP:conf/ijcai/SiL00LW21} and use a subset containing 9000 queries.

\subsection{Baselines}
For few-shot classification tasks, we compare the proposed method with the following strong baselines: 

\textbf{(1) NN} \cite{iclr/BaoWCB20} aims to train a classifier which assigns labels to unseen instances by utilizing the Euclidean distance metric to evaluate similarity with known data.

\textbf{(2) Prototypical Network} \cite{snell2017prototypical} is a metric-based few-shot method that calculates class prototypes by averaging support samples of each class. It predicts the category of query samples by employing the negative Euclidean distance between them and prototypes.

\textbf{(3) MAML} \cite{finn2017model} is an optimization-based method that learns an effective model initialization, which aims to utilize a small number of gradient updates and adapts the model to new tasks. 

\textbf{(4) Induction Network} \cite{DBLP:conf/emnlp/GengLLZJS19} acquires the generalized class-wise representations through the dynamic routing algorithm. It demonstrates promising performance in few-shot classification tasks by effectively capturing the relationships between samples and classes.

\textbf{(5) TPN} \cite{tpn} effectively leverages the information from both supports and queries through episodic training and a specialized graph construction, which can transfer labels from support samples to query samples. 

\textbf{(6) DS-FSL} \cite{iclr/BaoWCB20} is a meta-learning method which integrates the distributional signatures into attention scores. This approach effectively utilizes the distributional information to enhance the adaptability to new classes in few-shot text classification tasks.

\textbf{(7) MLADA} \cite{mlada} incorporates an adversarial domain adaptation network into a meta-learning framework, aiming to improve adaptability for new tasks and generate high-quality sentence embeddings.

\textbf{(8) ContrastNet} \cite{ContrastNet} is a supervised contrastive learning model to avert the task-level and instance-level  overfitting issues through the incorporation of two unsupervised contrastive losses.

For zero-shot tasks, we follow \cite{DBLP:conf/ijcai/SiL00LW21} and adopt the transformation-based methods with a linear classifier and use CNN, LSTM and BERT as the feature extractor, and the other strong baselines are as follows.

\textbf{(9) CDSSM} \cite{chen2016zero} uses a convolutional deep structured semantic model to create embeddings for intents that have not been seen during training.

\textbf{(10) ZSDNN} \cite{kumar2017zero} is an approach that enables zero-shot capabilities in intent detection by leveraging a common high-dimensional embedding space. 

\textbf{(11) CapsNet} \cite{xia2018zero} is an innovative model based on capsule networks that effectively adapts to new intents through knowledge transfer.

\textbf{(12) CTIR} \cite{DBLP:conf/ijcai/SiL00LW21} predicts unseen intents using their label names as input, guided by a multi-task learning objective to enhance intent discrimination, and employs a similarity scorer for more accurate intent connections.

\begin{table}[!t]
    \small
    \centering
    \renewcommand\arraystretch{0.9}{
    \begin{tabular}{l|cc|cc}
	\toprule
        {\textbf{Method}} & \multicolumn{2}{c|}{\textbf{SNIPS}} & \multicolumn{2}{c}{\textbf{CLINC}}\\
        \cmidrule(lr){2-5}
	{} & Acc & F1 & Acc & F1\\
	\midrule
	LSTM & $0.7947$ & $0.7918$ & $0.7173$ & $0.6873$ \\
	CNN  & $0.6515$ & $0.6091$ & $0.7303$ & $0.7094$ \\
	BERT & $0.7366$ & $0.7332$ & $0.6273$ & $0.5945$ \\
	CDSSM  & $0.6898$ & $0.6652$ & $0.6480$ & $0.6114$ \\
        ZSDNN & $0.8096$ & $0.8074$ & $0.8220$ & $0.8218$ \\
        CapsNet & $0.7421$ & $0.7258$ & $0.6451$ & $0.6187$ \\
        CTIR-BERT  & $\bf{0.9613}$ & $\bf{0.9612}$ & $0.8872$ & $0.8845$ \\
        CTIR-ZSDNN & $0.9507$ &$ 0.9507$ & $0.9357$ & $0.9362$  \\
        CTIR-CapsNet & $0.9484$ & $0.9484$ & $0.8701$ & $0.8691$ \\
        \midrule
        \textbf{Ours-top} & $\bf{0.9625}$ & $\bf{0.9624}$ & $\bf{0.9787}$ & $\bf{0.9786}$ \\
        \textbf{Ours-avg} & $0.9426$ & $0.9424$ & $\bf{0.9785}$ & $\bf{0.9784}$ \\
	\bottomrule
    \end{tabular}}
    \caption{Experimental results of zero-shot classification.}
    \label{tab:result2}
\end{table}

\subsection{Evaluation Metrics}

We evaluate the classification performance by using two widely adopted metrics: accuracy (Acc) and weighted-average F1-score (F1). These metrics are calculated by averaging the values weighted by the sample ratio of the corresponding class. 

\subsection{Implementation Details}

\subsubsection{Data Splitting}
For the few-shot tasks (including the 5-way 0-shot task), we perform a five-fold class split for each dataset, following the approach outlined in \cite{ContrastNet}. The training and validation class sets are exclusively utilized for training the baseline models, which serve as a reference for comparison. The evaluation of our method is conducted solely on the test set, simulating real-world scenarios where only unseen data is available during inference. To comprehensively assess our model's effectiveness, we perform evaluations on 10 distinct test tasks randomly sampled from the test set for each data partition. Within each task, 25 samples from each of the unseen classes are randomly selected to serve as test instances (as 25 queries).

For the zero-shot learning tasks, we adopt the experimental data partitioning approach proposed in \cite{DBLP:conf/ijcai/SiL00LW21}. The dataset is divided into seen classes and unseen classes, where the seen classes are only used for training the baseline models. And our proposed method is solely tested on the unseen classes. 

\setlength{\tabcolsep}{2mm}{
    \begin{table}[t]
    \small
    \centering
    \renewcommand\arraystretch{1}{
        \begin{tabular}{l|l}
        \toprule
        \multicolumn{1}{l|}{\textbf{Dataset}} & \multicolumn{1}{c}{\textbf{Template}} \\
        \midrule
  	\textbf{20News} & Here is news about \underline{ \scriptsize{\textless Category  Description}\textgreater }: `` \\
        \textbf{Amazon} & Here is a product review of  \underline{ \scriptsize{\textless Category  Description}\textgreater} :``\\
	\textbf{HuffPost} & A news headline about \underline{ \scriptsize{\textless Category  Description}\textgreater}:`` \\
	\textbf{Reuters} & Here is news about \underline{ \scriptsize{\textless Category  Description}\textgreater}:`` \\
	\textbf{SNIPS} & If I want to \underline{ \scriptsize{\textless Category  Description}\textgreater},  I will say ``\\
	\textbf{CLINC} & If I want to \underline{ \scriptsize{\textless Category  Description}\textgreater},  I will say ``\\
	\bottomrule
	\end{tabular}}
    \caption{The designed templates.}
    \label{tab:template}
\end{table}
}
 
\subsubsection{Parameter Settings}
For the anchor generation stage, we employ GPT2-XL as the generative pre-trained language model and adopt the top-p and top-k methods \cite{DBLP:conf/iclr/HoltzmanBDFC20,DBLP:conf/acl/LewisDF18,DBLP:conf/acl/ChoiBGHBF18} to sample generated words, which enhances the diversity of the generated data. Specifically, we set the parameter $p = 0.9$ and $k = 40$. For each class, we generate a maximum of 1000 samples, and each sample contains up to 100 tokens for datasets like Amazon, Reuters, and 20News, and up to 30 tokens for datasets like HuffPost, SNIPS, and CLINC. And the manual generation templates we use are shown in Table \ref{tab:template}. 
For the news and review datasets, we set $P=20$, while for the intent datasets, we set $P=60$, this is because we need to extract more information from the shorter utterances in the intent datasets.

For the classification reframing stage, we use the AdamW \cite{DBLP:journals/corr/abs-1711-05101} optimizer with a weight decay coefficient of 0.01 and apply label smoothing \cite{DBLP:conf/cvpr/SzegedyVISW16} to train the classifiers. We set the learning rate to 2e-5 for news and review datasets and 5e-5 for intent datasets, and use a linear warmup learning-rate with a proportion of 0.1 \cite{DBLP:journals/corr/abs-1907-11692}. For each task, we sample 100 anchor pairs as the validation set, and use the early stop training when the validation accuracy is 1 lasting for 3 epochs. 
To ensure the reliability and robustness of our results, all reported results are from 5 different runs.

\begin{table*}[!t]
\tabcolsep=0.17cm
\small
\centering
\renewcommand\arraystretch{1}{
    \setlength{\tabcolsep}{3.5mm}{
        \begin{tabular}{l|cc|cc|cc|cc}
	\toprule
	{\textbf{Method }} & \multicolumn{2}{c|}{\textbf{20News}} & \multicolumn{2}{c|}{\textbf{Amazon}}  & \multicolumn{2}{c|}{\textbf{HuffPost}} & \multicolumn{2}{c}{\textbf{Reuters}}\\
	\cmidrule(lr){2-3} \cmidrule(lr){4-5} \cmidrule(lr){6-7} \cmidrule(lr){8-9} 
	{}  & Acc & F1 & Acc & F1 & Acc & F1 & Acc & F1\\
	\midrule
        TPN (1-shot) & $0.5898$ & $0.5247$ & $0.7072$ & $0.6745$ & $0.5005$ & $0.4780$ & $0.8368$ & $0.8022$ \\
        DS-FSL (1-shot) & $0.6019$ & $0.5729$  & $0.6592$ & $0.6434$ & $0.3946$ & $0.3759$ & $0.7572$ & $0.7447$ \\
        MLADA (1-shot) & $0.6040$ & $0.5776$ & $0.6328$ & $0.6143$ & $0.4433$ & $0.4188$ & $0.7178$ & $0.7028$ \\ 
        ContrastNet (1-shot) & $\bf{0.7229}$ & $\bf{0.7047}$ & $\bf{0.7606}$ & $\bf{0.7548}$ & $0.5045$ & $0.4909$ & $0.8787$ & $\bf{0.8699}$ \\
        \midrule 
        \textbf{Ours-top (0-shot)} & $\bf{0.6875}$ & $\bf{0.6802}$ & $0.7450$ & $0.7319$ & $\bf{0.6837}$ & $\bf{0.6811}$ & $\bf{0.8830}$ & $0.8680$ \\
        \textbf{Ours-avg (0-shot)} &$0.6834$ & $0.6758$ & $\bf{0.7552}$ & $\bf{0.7444}$ & $\bf{0.6938}$ & $\bf{0.6911}$ & $\bf{0.8830}$ & $\bf{0.8695}$ \\
	\bottomrule
	\end{tabular}}}
    \caption{Experimental results of zero-shot classification (ours) comparing with 1-shot classification (baselines).}
    \label{tab:result3}
\end{table*}

\begin{table}[t]
\small
\centering
\renewcommand\arraystretch{1}{
    \begin{tabular}{l|cc|cc}
    \toprule
    {\textbf{Method}} & \multicolumn{2}{c|}{\textbf{SNIPS}} & \multicolumn{2}{c}{\textbf{CLINC}} \\
    \cmidrule(lr){2-5}
    {} & Acc & F1 & Acc & F1\\
    \midrule
    SL & $0.8360$ & $0.8322$ & $0.9510$ &$0.9506$ \\
    GPT-3.5 & $0.9200$ & $0.9203$ & $0.9500$ & $0.9467$ \\
    \midrule
    \textbf{Ours-top} & $\bf{0.9500}$ & $\bf{0.9502}$ & $\bf{0.9880}$ & $\bf{0.9879}$ \\
    \textbf{Ours-avg} & $\bf{0.9270}$ & $\bf{0.9272}$ & $\bf{0.9840}$ & $\bf{0.9838}$ \\
    \bottomrule
    \end{tabular}}
    \caption{Comparison with the supervised learning (SL) classifier trained with the generated data and GPT-3.5.}
    \label{tab:result4}
\end{table}

\subsection{Experimental Results}
\subsubsection{Comparison with Baselines}
The experiment results are reported in Table \ref{tab:result1} and Table \ref{tab:result2}, where 
\textbf{ours-top} and \textbf{ours-avg} represent the experiment results predicted by the top-one score and the average score respectively. Some baseline results are taken from \cite{DBLP:conf/ijcai/SiL00LW21}. And the top-2 results are highlighted in bold. Based on the results, we can make the following observations. 

(1) For few-shot text classification, our model outperforms other strong baselines in most cases. This is because our model can introduce useful knowledge of unseen classes by leveraging the anchor generation procedure and reduce the search space by reframing the multi-class classification task as the binary classification task, thus improving the performance significantly.

(2) The zero-shot comparison experiment shows that our model significantly outperforms the other zero-shot strong baselines on CLINC. For the SNIPS dataset, our approach also has a competitive result, but the improvement is limited. The reason is that the SNIPS dataset is only separated into two unseen classes, which makes the advantages of classification reframing can not be fully realized.

(3) Our model can transfer smoothly between few-shot and zero-shot by controlling the composition of anchors. We also report the performance of our model without supports ($N$-way $0$-shot) in Table \ref{tab:result3}, which shows that our model is still competitive in $0$-shot setting compared to the baseline in $1$-shot setting. That is because the generated data effectively make up for the lack of support sets. This further proves the superiority of data generation for novel classes in the case of insufficient resources, and the improvement effect for text classification tasks of screening out category anchors and classification reframing.

\subsubsection{Comparison with Supervised Learning and LLMs}
In recent times, large-scale language models (LLMs) such as GPT-3.5 have exhibited remarkable capabilities in zero-shot tasks. Moreover, a clear solution to tackle this challenge is to employ a conventional multi-classifier supervised trained on the abundant generated data from the anchor generation stage. As an extension of our research, we conduct supplementary experiments in the zero-shot intent detection domain. We present the experimental outcomes in Table \ref{tab:result4}, showcasing the results of randomly selecting 200 samples for testing. A comprehensive comparison is made among our method, the utilization of GPT-3.5, and a traditional supervised learning multi-classifier approach based on BERT. Based on the results, our method outperforms emerging large-scale models and traditional classifiers reliant on substantial volumes of generated data. This advantage primarily stems from the extraction of generated data in the anchor screening stage, coupled with the enhanced prediction efficiency of the model facilitated by the classification reframing module. This combination refines the model's focus on assimilating the implicit distinctions among categories.

\subsubsection{Performance with Different Number of Anchors}
\begin{figure}[t]
\centering
\includegraphics[width=0.48\textwidth]{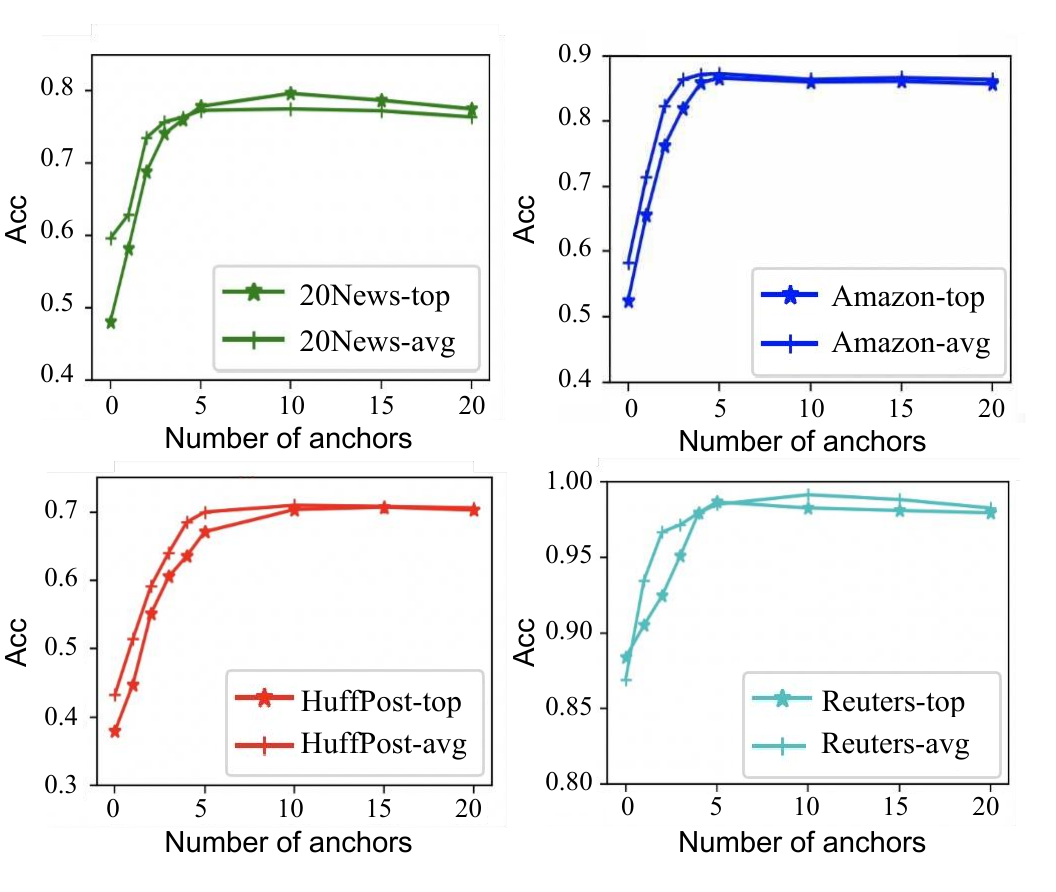}
\caption{Comparison of the classification accuracy across different anchor count settings.}
\label{holdnum}
\end{figure}

Figure \ref{holdnum} depicts the impact of anchor count per category on classification accuracy under the 5-way 5-shot setting for a single data split. In the news and review datasets, as the number of anchors increases, there is a progressive enhancement in recognition accuracy, with the improvement eventually leveling off and reaching stable performance.

\section{Conclusion}
In this paper, we propose a simple yet effective framework for addressing both zero-shot and few-shot text classification tasks without any seen classes. By employing anchor generation, our approach effectively sidesteps negative transfer from seen classes and accurately models the unseen categories by their own descriptions. By reframing the intricate multi-class classification task into a more feasible binary-classification task, our model can improve the classification performance effectively. Extensive experimental results on widely-used public datasets confirm the superiority of our method over other strong baselines. In future work, we plan to extend the proposed framework to deal with multi-label few-shot and zero-shot text classification tasks.

\section{Acknowledgments}
The authors are grateful to the reviewers for their valuable comments. This work was supported by National Natural Science Foundation of China (No. 62106035, 62206038, 61972065) and Fundamental Research Funds for the Central Universities (No. DUT20RC(3)040, DUT20RC(3)066), and supported in part by Key Research Project of Zhejiang Lab (No. 2022PI0AC01), National Key Research and Development Program of China (2022YFB4500300). We also would like to thank Dalian Ascend AI Computing Center and Dalian Ascend AI Ecosystem Innovation Center for providing inclusive computing power and technical support.

\bibliography{aaai24}

\end{document}